\pdfoutput=1
\documentclass[11pt]{article}

\usepackage{ACL2023}

\usepackage{times}
\usepackage{latexsym}
\usepackage[T1]{fontenc}
\usepackage[utf8]{inputenc}

\usepackage{hyperref}
\usepackage{microtype}

\usepackage{inconsolata}

\title{On the Use of Metaphor Translation in Psychiatry
}

\author{Lois Wong \\
  Computer Science Department \\ 
  Johns Hopkins University \\
    \texttt{lwong23@jhu.edu } \\}

\begin{document}
{\makeatletter\acl@finalcopytrue
  \maketitle
}

\begin{abstract}
Providing mental healthcare to individuals with limited English proficiency (LEP) remains a pressing problem within psychiatry. Because the majority of individuals trained in providing psychiatric care are English speakers, the quality of mental healthcare given to LEP patients is significantly lower than that provided for English speakers. The provision of mental healthcare is contingent on communication and understanding between the patient and healthcare provider, much more so than in the realm of physical healthcare, and English speakers are often unable to comprehend figurative language such as metaphors used by LEPs. Hence, Figurative Language Translation is invaluable to providing equitable psychiatric care. Now, metaphor has been shown to be paramount in both identifying individuals struggling with mental problems and helping those individuals understand and communicate their experiences. Therefore, this paper aims to survey the potential of Machine Translation for providing equitable psychiatric healthcare and highlights the need for further research on the transferability of existing machine and metaphor translation research in the domain of psychiatry. 
\end{abstract}

\section{Introduction}

One of the most neglected areas of psychiatry is providing mental healthcare to non-English speakers. The healthcare quality given to psychiatric patients with limited English proficiency (LEP) is markedly lower than that provided for English-speaking patients, and this is due in large part to the fact that the majority of individuals qualified to provide such help are English speakers. English speakers often demonstrate an inability to comprehend figurative language such as metaphors used by LEP patients and hence are unable to provide equitable care. Figurative Language Translation is therefore an extremely important sub-problem that must be solved in order to achieve the broader goal of providing quality psychiatric support to LEP individuals. The following definition of metaphor effectively illustrates the importance of metaphor in mental health:

\begin{quote}
“From the perspective of cognition, the essence of a metaphor is the understanding and experiencing of one kind of thing in terms of another. [...] To alleviate the lack of appropriate words to express new concepts in a certain period of time, humans create new metaphorical meanings of words using active imagination” \cite{Chen:2021}.

\end{quote}

Metaphor has been shown to be a device used by depressed individuals to make sense of and describe their experience—the comparison between “uncertain and unfamiliar experiences to familiar references helps people communicate complicated, sensitive, and emotional topics” \cite{Shi:2023}; it is “a pivotal tool for helping people with depression better understand themselves and their problems [...], facilitating effective communication between therapists and patients” \cite{Han:2022}. Additionally, the automatic detection of mental illness from social media posts using metaphorical mappings has been proven to be tremendously effective, helping users “privately and conveniently understand their mental health status in the early stages before seeing mental health professionals \cite{Han:2022}. Hence, from a brief survey of the applications of machine translation in the provision of mental healthcare, we summarize the open research challenges in the areas of Metaphor Identification, Metaphor Paraphrasing, and model explainability, and highlight the need for further research in the transferability and adoption of existing machine and metaphor translation research in the realm of psychiatry.

\section{Human interpretation and why it comes short}

As of now, the usage of human interpreters is shown to be effective in the provision of general healthcare to LEP individuals. Jacobs et al. assess the practicality and viability of using interpreters for providing healthcare services to LEP Spanish-speaking patients in “Overcoming Language Barriers in Health Care: Costs and Benefits of Interpreter Services.” They measured changes in delivery and cost for patients with and without interpreter services and found that the patients with interpreters “received significantly more recommended preventive services, made more office visits, and had more prescriptions written and filled” \cite{Jacobs:2004} than those without.

While the use of human interpreters is regarded to be the gold standard of providing both physical and mental healthcare for LEPs, this solution is rarely employed in the context of psychiatry, with “less than 20\% of clinical encounters [using] interpreting services [...] due to time constraints” \cite{Tougas:2022}. In fact, Tougas et al. argue that human interpretation is the key barrier to providing quality mental healthcare to LEP patients, as it renders “a less comprehensive picture of the patient’s psychopathology” for two reasons: interpretation guidelines are designed to assist the provider (rather than the patient) and patients are inclined, if not advised, to slow and simplify their speech for the translator’s benefit.

\section{Asynchronous Telepsychiatry}

Tougas et al. examine methods for improving the mental healthcare quality for patients with limited English proficiency in “The Use of Automated Machine Translation to Translate Figurative Language in a Clinical Setting: Analysis of a Convenience Sample of Patients Drawn From a Randomized Controlled Trial”. They note that using human interpreters “has been associated with a reduced number of follow-up appointments, reduced patient and provider satisfaction, and an increased likelihood of not asking the questions that the patient wanted to ask” \cite{Tougas:2022} and believe asynchronous telepsychiatry to be a promising solution to this problem. Asynchronous telepsychiatry (ATP) is a treatment format where a psychiatric patient’s answers to clinical questions are recorded, transcribed, translated, and shown to a psychiatrist (as captions to a video recording) to evaluate and prescribe a treatment plan for. ATP is regarded as a clinically valid method for psychiatric assessment, and Tougas et al. believe that automated speech recognition and automated machine translation can significantly augment the current use and effectiveness thereof. To approximate patient speech complexity and quantity in clinical encounters requiring interpretation, the authors measured “the frequency and accuracy of figurative language device (FLD) translations and patient word count per minute in psychiatric interviews” \cite{Tougas:2022} for 6 Spanish-speaking patients. These patients underwent two assessments: the first conducted by an English-speaking psychiatrist and interpreter, and the second with a Spanish-speaking trained mental health interviewer-researcher and AI interpretation. The first option is representative of the current gold standard of LEP patient interviews while the second employs the ATP App, a dual automated speech recognition and machine translation function, to represent the new asynchronous telepsychiatric format.

The results of this experiment indicate that interviews conducted in Spanish with AI interpretation individuals boast a significant increase in figurative language device (FLD) use and patient word count per minute, whereas real-time consultations with a psychiatrist and translator tend to consist of simplified and shortened patient speech. Additionally, both human and AI translations of FLDs were found to be somewhat inaccurate, finding an aggregate translational accuracy of 52\% (human) vs 30\% (AI), (P > 0.5) \cite{Tougas:2022}. Interestingly, the authors noticed that much of the machine translation inaccuracy comes from erroneous transcriptions, suggesting that improved audio and transcription quality could boost the accuracy rate of the machine translations. Tougas et al. demonstrate the potential of asynchronous telepsychiatric care, a more convenient method that also increases expressive freedom for the patient, and highlights the need for further research and development in the areas of machine translation for figurative language and asynchronous telepsychiatry.

\section{Metaphor in mental health management}

Shi et al. confirm the value of examining metaphor usage in mental healthcare in “Words for the Hearts: A Corpus Study of Metaphors in Online Depression Communities.” Drawing off Lakoff and Johnson’s Conceptual Metaphor Theory (CMT) which posits metaphor “as a cognitive tool for understanding and organizing experience” \cite{Shi:2023}, the authors believe that insight into how individuals use metaphor to make sense of and describe their experience can not only make counseling and interventions more effective but also “help identify and manage mental health issues'' \cite{Shi:2023}. Using a web crawler to compile a corpus from a Chinese online depression community, Shi et al. analyze the metaphors used in online health communities (OHC). They extracted 652 metaphorical expressions and categorized them as “personal life”, “interpersonal relationship”, “time”, or “cyber-culture”.

Interestingly, self-conceptualization metaphors were found to be a distinctive feature of depressed individuals, which Shi et al. connected to prior research showing first-person singular pronouns and death/negative emotion words to be characteristic of depression. Additionally, “container” metaphors were found to be prevalent in that “people with depression are conceptualized as “containers of negative emotions and confined in enclosed spaces from which they cannot escape” \cite{Shi:2023}. Citing several other insights drawn from their study, the authors show that “metaphors can be a very powerful tool in healthcare” \cite{Shi:2023}. Not only can depressed individuals express their pain through metaphor, but identifying and understanding these people can also be achieved by decoding them.

Similarly, Han et al. show metaphor to be a valuable tool for online mental health management in “Hierarchical Attention Network for Explainable Depression Detection on Twitter Aided by Metaphor Concept Mappings.” Reiterating how depressed individuals use metaphor to describe their experience and emotions, the authors leverage metaphorical concept mappings (MCM) as inputs to their explainable model for detecting depression among Twitter users. MCMs are represented as a “sequence of A IS B”, where A is a target concept, B is a source concept, and IS a relation mapping A to B” \cite{Han:2022}; the user is hence represented as a set of tweets and a set of metaphorical mappings in their tweets. Han et al.’s three-step process of acquiring concept mappings for their model is identifying the metaphor, paraphrasing the metaphor, and generating concept mappings \cite{Han:2022}, and the authors synthesize pre-existing algorithms for these tasks.

The resulting Hierarchical Attention Network (HAN) model is a binary classifier that outperforms the strongest baseline (Zogan et al., 2021) with an increased 6.0\% F1 score on average and improves upon the most competitive benchmark encoder for their task (Hochreiter and Schmidhuber’s LSTM (1997)) by 1.9\% on a validation set while using only a quarter of LSTM’s parameters. Using context-level attention mechanisms, the model finds the relative importance of certain tweets and metaphors, and offers insight into the cognition of depressed individuals. As such, HAN “not only detects depressed individuals but also identifies features of such users’ tweets and associated metaphor concept mappings.” \cite{Han:2022}

Han et al. are the first to introduce MCMs as a feature to “improve explainability and performance” \cite{Han:2022}. They use attention weights to visualize and justify their model’s decision-making process, quantitatively describing “how much each tweet and MCM contributes to a predicted label” \cite{Han:2022}. Achieving an average F1 score of 97.2%
and an accuracy of 97.1\% \cite{Han:2022}, the authors prove the effectiveness of using metaphor to detect depression.

\section{Metaphor translation is hard}

Thus far, metaphor has been shown to be a valuable tool that can not only make psychiatric care more effective but also “help identify and manage mental health issues” \cite{Shi:2023}. However, figurative language translation is a difficult endeavor. Forner provides theoretical and historical insight into why machines struggle with translating natural language, particularly metaphor, in Language @t work: Language Learning, Discourse and Translation Studies in Internet. Noting the failure of the 1950s’ formal linguistic approaches to Machine Translation (MT) due to the sheer complexity of linguistic problems, he reiterates the sentiment that computer systems will “never reach the knowledge of the world that human beings have'' \cite{Forner:2005}. This period of disillusionment ended in the late 1980s when interlingual systems, translation models based on “abstract language-neutral representations”, suggested that AI can be used to improve MT systems. The key barrier to this solution, however, lies in the non-compositional nature of metaphor—the meaning of an utterance is more than the composition of its parts. Since computers “can only operate with a set of formal instructions” \cite{Forner:2005}, logic cannot always be used to code/decode language.

\section{Key research challenges in metaphor translation}

We partition the endeavor of metaphor translation into two key research challenges: metaphor identification and metaphor understanding, which is achieved by metaphor paraphrasing; both of these methods have been almost exclusively studied outside the context of mental health.

\subsection{Metaphor Identification}

Shi et al. examine and illustrate why metaphor identification is difficult in Chinese while conducting their study of Metaphors in Online Depression Communities. Chinese uses a monosyllabic writing system, unlike English, with each character corresponding to a single phonological syllable. However, a word in Chinese can contain multiple characters without explicit word delimiters, and their simple word forms and commonality of compound words make it very difficult to distinguish word classes (parts of speech) due to Chinese's minimal morphology. As such, it is often challenging to “delineate units of analysis” \cite{Shi:2023}.

As such, Shi et al. address the problem of metaphor identification and attempt to develop a more reliable and efficient procedure for metaphor identification in Chinese corpora by combining bottom-up (collect metaphors as you process text) and top-down (start with a list of known metaphors) approaches. The authors proposed a procedure for Chinese metaphor identification where the workflow is to “ (1) combine MIPVU to identify metaphorical expressions (ME) bottom-up and formulate preliminary working hypotheses, (2) collect more ME top-down in the corpus by performing semantic domain analysis on identified ME, [and] (3) analyze ME and categorize conceptual metaphors using a reference list” \cite{Shi:2023}. Through this method, the authors were additionally able to identify the sociocultural environment that depressed individuals are experiencing as the “lack of offline support, social stigmatization, and substitutability of offline support with online support” \cite{Shi:2023}.

Metaphor identification is further studied by Chen et al. in “Metaphor identification: A contextual inconsistency based neural sequence labeling approach.” In their paper, the authors hypothesized that a greater semantic inconsistency between a target word and its contextual words leads to an increased likelihood of that target word being a metaphor. For example, consider the sentences “The painting won critical acclaim,” vs “This army won a battle.”—the first sentence is metaphorical while the second is not. The authors note that the categories of “army” and “won” are identical, i.e.  \textit{battle}  while that of “painting” belongs to a different category,  \textit{art} \cite{Chen:2021}. They use this property of “semantic contextual inconsistency” to formulate metaphor identification as a “sequential tagging problem” \cite{Chen:2021}. As such, the authors represent a collection of sentences C as a set of each sentence’s words and a corresponding set of the sentence’s word labels, i.e. metaphorical or not metaphorical. They train a model on C to predict the label of each word for unknown sentences; the semantic meaning of each word is then represented with an abstract distributed representation and the distributional distance between each word pair in a sentence is used to measure contextual inconsistency.

From this, Chen et al. developed SEQ-CI, the first neural sequence labeling model that uses contextual inconsistency to identify metaphors. This model consists of five main components: “the embedding layer, the contextualized word representation layer, the abstract distributed representation layer, the semantic contextual inconsistency module, and the metaphor prediction layer” \cite{Chen:2021}. The authors use the Bi-LSTM RNN model to learn contextualized word representations and also as a sequence encoder, and their abstract distributed representation layer uses a softmax function “which takes the contextualized representations as input and generates corresponding distributed representations of individual words” \cite{Chen:2021}. Then, the metaphorical layer takes the contextualized representations as input for the sequential classification task. The results of SEQ-CI are promising, and the authors show that it performs well over the seven well-established baselines of Semantic-Feature, Abstractness, Word-Similarity, $CNN+RNN_{ensmb}$, $RNN\_ELMo_{SEQ}$, RNN\_HG, and RNN\_MHCA.

\subsection{Metaphor Paraphrasing}

Now, metaphor identification is only the first step to translating metaphor. After a metaphor is identified in some text, the next challenge is to understand or decode it. In “Interpreting Verbal Metaphors by Paraphrasing,” Mao et al. experiments with interpreting verbal metaphors since verbs are the most common metaphorical part-of-speech category. Their proposed method of using BERT and WordNet hypernyms and synonyms to interpret metaphors significantly outperforms the current state-of-the-art baseline and is shown to help an MT system improve metaphor translation accuracy from English to 8 target languages.

Mao et al. begin by defining metaphor interpretation as a paraphrasing task, i.e. predicting a metaphor’s literal counterpart. To do this, they use BERT to predict a sequence of words that is most similar to a metaphor’s intended meaning, or the word(s) that has the highest probability of appearing as “the literal paraphrase of the metaphor” \cite{Mao:2021}. For sentences containing several metaphors, the authors mask one metaphoric word at each time. This method is fully unsupervised since they use WordNet to constrain the semantically similar words for the masked prediction. This proved to be very effective: Mao et al.’s model achieves an 8\% accuracy gain on the state-of-the-art baseline for verbal metaphor paraphrasing and 13\% gain on two benchmark datasets for human-evaluated paraphrasing tasks,
MOH and VUA; in addition, their method increases the accuracy of English metaphor translation by 20.9\% \cite{Mao:2021}.

Subsequently, Qiang et al. propose Chinese Idiom Paraphrasing (CIP) as a novel pre-processing technique to improve machine translation system performance in “Chinese Idiom Paraphrasing.” Treating CIP as a paraphrase generation task, CIP rephrases sentences containing idioms as non-idiomatic sentences while preserving the original sentence’s meaning. Given a source sentence containing at least one idiom, the task of CIP is to output a target sentence that contains no idioms but still preserves the original sentence’s meaning. The authors build a dataset for CIP by designing a framework that constructs a pseudo-CIP dataset and asking humans to revise and evaluate it \cite{Qiang:2022}. They then deploy five baselines for rephrasing the input idiom-included sentence, experimenting with three Seq2Seq models: LSTM-based, Transformer-based, and mT5-based, and propose two new CIP methods, Infill-based and Knowledge-based. The knowledge-based CIP method is an extension of the Seq2Seq model that allows idioms to be replaced “with their representations at the same locations in source sentences” \cite{Qiang:2022}, which in turn allows the network to better leverage idioms. A sentence is then encoded by extracting a dictionary of idiom interpretations from the authors’ training dataset and concatenating original sentences with every idiom interpretation that is used for training their model. The Infill-based method generates the target sentence from the original idiomatic sentence based on context; this method leverages span-masked language modeling (MLM) which “reconstructs consecutive spans of input tokens and masks them with a mask token” \cite{Qiang:2022} with the help of mT5. Essentially, this method reconstructs the idiom’s interpretation by replacing the idiom with the mask token.

The results of these five models are promising, outperforming the two unsupervised models of Re-translation and BERT in five metrics. For in-domain performance, the Infill-based model has a BLEU score of 84.88 and the Knowledge-based model scored 85.25, significantly higher than BERT’s score of 73.58. Out-domain wise, the Infill model scored 87.19 and the Knowledge model scored 85.25 while BERT scored 79.65. For human evaluations of simplicity, meaning, and fluency, the Infill model scored 3.96 in-domain and 3.76 out-domain and the Knowledge model scored 3.77 in-domain and 3.75 out-domain, while BERT scored 2.72 in-domain and 2.31 out-domain, thereby establishing a large-scale benchmark for CIP.

\section{Conclusion}

Since most of the research on metaphor translation was conducted outside the domain of psychiatry, it is worth looking into the integration of machine translation into existing psychiatric practice. In “Modeling workflow to design machine translation applications for public health practice,” Turner et al. seek to understand the information workflow process used to translate public health materials to “inform the design of context-driven machine translation tools for public health \cite{Turner:2014}. While conducting interviews, performing a task analysis, and validating results with public health professionals, the authors modeled a “translation workflow and [identified] functional requirements for a translation system for public health” \cite{Turner:2014}. They found that the key processes that prevented multilingual health materials in public health include the lack of standardization across departments, as well as time and cost. In addition, quality assurance was a large concern, particularly for lower-profile documents such as letters and notes. Noting the need for human quality assurance, the authors found that using MT for more than a few words or phrases came with strong reservations for many people—evidence that MT can deliver a quality end product is needed. In addition, Turner et al. highlight the necessity for “a clear idea of how MT can fit into [the] current translation workflow” \cite{Turner:2014}. They also state that while MT can be used to “facilitate and accelerate the production of multilingual health materials” \cite{Turner:2014}, it “always needs to be followed by human review and corrections,” proposing a widely-used approach in the commercial sector known as MT plus post-editing.

In order to adapt MT methods for providing equitable psychiatric care for LEP patients, further research is additionally needed in the integration of MT in psychiatry since many of the current methods for machine and metaphor translation are outside the domain of mental health. On a somewhat related note, there is a need for further research on the transferability of the existing methods of metaphor translation into psychiatry. It would also be beneficial to investigate further into leveraging these methods to improve Tougas’ proposed framework of asynchronous telepsychiatry. Finally, additional research in developing explainable and accurate models is in high demand, and would be instrumental to assuaging the distrust of using Machine Translation in healthcare.

%%%%

\bibliographystyle{acl_natbib}
\bibliography{custom}

\end{document}